\documentclass[journal]{IEEEtai}

\usepackage{fancyhdr}
\usepackage{extramarks}
\usepackage{graphicx}

\usepackage{amsthm}
\usepackage{amsmath}
\usepackage{amsfonts}
\usepackage{comment}
\usepackage{csquotes}
\usepackage{xspace}
\usepackage{siunitx}

\usepackage{algorithm}
\usepackage[normalem]{ulem}

\usepackage{color,array}

\usepackage{todonotes}

\usepackage{amsmath} 
\usepackage{tikz}
\newcommand{\eps}{\epsilon}

\usepackage{float}

\usepackage[ruled,algo2e]{algorithm2e}
\usepackage{amsfonts}
\usepackage{amssymb}
\usepackage{algpseudocode}
\usepackage[all]{xy}
\usepackage{tikz-cd}
\usepackage{multirow}
\usepackage{bm}
\usepackage{booktabs}
\usepackage[numbers,sort&compress]{natbib}
\usepackage{subfigure}
\usepackage{palatino}
\usepackage{caption}
\usepackage{subcaption}

\newtheorem{lemma}{Lemma}
\usepackage[colorlinks,linkcolor=black,anchorcolor=black,citecolor=black,urlcolor=blue]{hyperref}

\usepackage{booktabs}
\usepackage{mathtools}
\usepackage{amssymb}
\usepackage{capt-of}
\usepackage{mciteplus}
\usepackage{cite}
\usepackage{mathrsfs}
\usepackage{parskip}
\usepackage{soul}
\usepackage{textcomp}
\usepackage[colaction]{multicol}
\usepackage[switch]{lineno}
\usepackage{lipsum}
\usepackage{etoolbox}
\usepackage{longtable}
\usepackage{array}
\usepackage{tablefootnote}
\usepackage{ragged2e}
\usepackage{lscape}
\newcolumntype{C}[1]{>{\centering\arraybackslash}p{#1}}
\usetikzlibrary{automata,positioning}
\usetikzlibrary{automata,positioning}

\newtheorem{definition}{Definition}

\usepackage{tikz,xcolor,hyperref}
\definecolor{lime}{HTML}{A6CE39}
\DeclareRobustCommand{\orcidicon}{%
	\begin{tikzpicture}
	\draw[lime, fill=lime] (0,0)
	circle [radius=0.16]
	node[white] {{\fontfamily{qag}\selectfont \tiny ID}};
	\draw[white, fill=white] (-0.0625,0.095)
	circle [radius=0.007];
	\end{tikzpicture}
	\hspace{-2mm}
}
\foreach \x in {A, ..., Z}{%
	\expandafter\xdef\csname orcid\x\endcsname{\noexpand\href{https://orcid.org/\csname orcidauthor\x\endcsname}{\noexpand\orcidicon}}
}
 
\author{Gokul Bhusal$^1$, Delanna Do$^2$, Kevin Miller$^3$, Ekaterina Merkurjev$^{4}$\footnote{Corresponding author,
	Email:  merkurje@msu.edu} and  Stephan Wojtowytsch$^2$      \\
$^1$ Department of Mathematics,\\
University of California, Los Angeles, CA 90095, USA.\\
$^2$Department of Mathematics,\\ University of Pittsburgh, Pittsburgh, PA 15213, USA.\\
$^3$  Department of Mathematics, \\ Brigham Young University, UT 84602, USA. \\
$^4$ Department of Mathematics and CMSE, \\
Michigan State University, MI 48824, USA.\\

}
\date{}

\begin{document}
\bstctlcite{IEEEexample:BSTcontrol}

\title{Fermat ProbCover: Coverage-Based Active Learning on Fermat Distance Graphs 
}

\maketitle

\begin{abstract}
Active learning aims to reduce annotation cost by selecting the most informative examples for labeling. In the low-budget regime, where uncertainty-based methods often fail, coverage-based active learning methods such as ProbCover have been shown to be effective. However, these methods typically rely on Euclidean distance, which may not adequately capture the underlying geometry of data. In this article, we propose PC-Fermat, a coverage-based active learning method that replaces Euclidean coverage neighborhoods with neighborhoods defined by a Fermat distance, i.e., a data-adaptive distance that reflects the underlying distribution of the data points. To facilitate computation, we construct a weighted Euclidean neighborhood graph and use a truncated shortest-path algorithm to compute the Fermat coverage graph. We also establish sufficient conditions under which this construction exactly preserves the Fermat coverage computed on the complete graph. Finally, we evaluate PC-Fermat on eight classification datasets across varying labeling budgets, with particular emphasis on the very low-budget regime. Our experimental results demonstrate that PC-Fermat outperforms several existing active learning methods on the majority of the datasets.

\end{abstract}

\begin{IEEEImpStatement}
Sufficient amounts of labeled data are essential for achieving strong performance in machine learning tasks. However, obtaining high-quality labeled data is often expensive, time-consuming, and subject to practical or ethical constraints. These challenges are particularly evident in domains such as healthcare, biology, and remote sensing, where expert annotation is costly and data collection may involve privacy and safety considerations. Active learning has emerged as a promising approach for addressing limited labeling budgets by selecting the most informative samples for annotation. By reducing the number of samples that require labeling while preserving data diversity and representativeness, active learning can substantially improve annotation efficiency. The proposed algorithm advances this goal, and our experimental results demonstrate that it consistently improves the performance of {\color{black}supervised and} semi-supervised learning algorithms while requiring fewer labeled samples. These improvements can help lower annotation costs, facilitate the use of machine learning methods in data-scarce settings, and broaden the applicability in real-world applications.
\end{IEEEImpStatement}



\section{Introduction}
\label{Introduction}
\IEEEPARstart{T}{he} remarkable success of machine learning--particularly deep learning--has arguably been driven by the availability of large amounts of labeled data.  However, obtaining sufficient amounts of labeled data for real-world applications is often expensive, time-consuming, and subject to ethical and practical considerations \citep{chai2024mitigating, bhusal2024persistent}.  Active learning \citep{settles_active_2012} seeks to address this challenge by reducing the amount of labeled data required for effective model training. Specifically, given a fixed labeling budget of $b$ examples, active learning methods aim to select a subset of $b$ examples from an unlabeled dataset for annotation, with the goal of maximizing the predictive performance of a model trained on the resulting labeled data. 


We distinguish one-shot coreset-selection methods from adaptive batch-selection methods in active learning. One-shot, coreset-selection methods \citep{hardiman2025topology} identify the set of examples for labeling up front, \emph{prior to} any data labeling or model training. In contrast, adaptive batch-selection methods repeat a three-step cycle at each iteration: (i) train a model on the current labeled set, (ii) adaptively select unlabeled examples for labeling, and (iii) update the labeled set. In either setting, one can think of the active learning methods as defining a criterion (or acquisition function) that quantifies the utility of including a currently unlabeled example in the set to be labeled. {\color{black}Coreset selection methods are useful for batch-selection methods, at least as an initial step and for maintaining data diversity.}

Acquisition functions for active learning can generally be classified into uncertainty-based \citep{settles_active_2012,bertozzi2018uncertainty,qiao_uncertainty_2019,miller2023poisson,bhusal2025malady} and representation-based \citep{yehuda2022active, hacohen2022active} criteria. Within the adaptive, batch-mode setting of active learning, uncertainty-based methods prioritize samples for which the current model is least confident, allowing the classifier to refine its decision boundaries and to maximize the added value of new annotation. Representation-based criteria, on the other hand, seek to select samples from diverse regions of the data distribution, ensuring that all regions of the input space are adequately covered or ``represented'' in the labeled set. 

One of the key challenges in the practical application of active learning is to balance uncertainty-based and representation-based query strategies, a problem that is often viewed as an \textit{ exploration-exploitation trade-off.} Representation-based methods encourage exploration by selecting samples that provide broad coverage of the underlying data distribution, whereas uncertainty-based methods promote exploitation by querying samples near the model's decision boundary to refine its predictions. This balance is particularly important in the low-budget regime, where many uncertainty-based active learning strategies may fail to outperform random sampling \citep{yehuda2022active, hacohen2022active}. In particular, models trained on only a small number of labeled examples can produce unreliable uncertainty estimates, limiting the effectiveness of uncertainty-based selection. In contrast, coreset-selection methods prioritize representative sample selection and can achieve better performance in low-budget settings. These observations highlight the importance of prioritizing exploration over exploitation during the early stages of active learning, \textcolor{black}{analogous to early-stage reinforcement learning}.

\subsection{Our Approach}

We focus on one-shot coreset-selection methods for representation-based active learning to support effective learning under limited labeling budgets.
Among such methods, ProbCover~\citep{yehuda2022active} provides a metric-based approach to coreset selection that seeks to maximize dataset coverage by selecting samples that effectively cover the underlying data distribution. {\color{black}In more technical terms, ProbCover produces a \emph{coverage graph} of $\epsilon$-nearest neighbor distances to tabulate the extent to which a selected coreset ``covers'' the dataset.} Importantly, ProbCover does not require a particular distance metric; rather, it simply relies on a metric structure over the representation space. This observation motivates our approach: instead of modifying the ProbCover selection mechanism, we investigate how its performance can be improved by equipping it with a data-adaptive metric that better captures the underlying geometry of high-dimensional data. 

The effectiveness of such an acquisition function depends critically on accurately measuring relationships between samples in the underlying data geometry.  The quality of coreset selection depends on the distance metric used to characterize the data geometry. In high-dimensional spaces, conventional Euclidean distances may fail to capture the intrinsic low-dimensional structure of the data distribution, potentially leading to less representative sample selection. 

\textbf{To address this issue, we integrate data-adaptive, shortest-path metrics to more accurately capture the data geometry within a ProbCover approach for coreset selection.} The idea is to construct a weighted graph associated with the dataset $\mathcal{X}$ and then compute shortest-path distances on the graph to capture the underlying structure of $\mathcal{X}$.  One particular class of density-weighted path metrics is the Fermat distance, where the edge weights are calculated using a power of the Euclidean distance. Fermat distance has been successfully applied to a range of problems, including semi-supervised methods and clustering methods \citep{little2022balancing, little2020path, mckenzie2019power, vincent2003density,bijral2012semi} and high-dimensional signal processing \citep{zhang2021hyperspectral,manousidaki2024clustering}. 

\subsection{Contributions} \label{subsec:contributions}

The contributions of the paper are the following:
\begin{enumerate}
    \item   We propose PC-Fermat, a variant of ProbCover that replaces Euclidean coverage neighborhoods with Fermat distance-based neighborhoods. This provides a data-adaptive approach to constructing coverage graphs that accounts for the underlying geometry of the data.
  
    \item We present an efficient procedure for constructing Fermat coverage graphs using a weighted Euclidean neighborhood graph and truncated shortest-path computations. We establish conditions under which this construction exactly preserves the Fermat neighborhoods defined on the complete graph.
      \item   We evaluate PC-Fermat across eight classification datasets and under varying labeling budgets. We compare its performance against ProbCover and several active learning methods.
\end{enumerate}

\vspace{0.2cm}
The remainder of the article is organized as follows. In Section \ref{sec: background}, we present background information on Fermat distance and graph construction techniques. In Section \ref{sec: proposed_method}, we derive our proposed method. The results of the experiments on benchmark datasets and the discussion of the results are presented in Section \ref{sec: results}. We gather some concluding remarks in Section \ref{sec: conclusion}.

\section{BACKGROUND}
\label{sec: background}
In this section, we introduce the $p$-Fermat distance,
which forms the basis of our proposed method. We then
describe the construction of the Fermat coverage graph
and present an efficient procedure for computing its
neighborhoods. The notation used throughout the paper
is summarized in Table~\ref{tab:notations}.

\begin{table}[!ht]
    \centering
    \vspace{0.05cm}
    \begin{tabular}{ll}
        \hline
        Variable & Description\\
        \hline
      
        $\mathcal{X}$ & dataset, vertex set (both data and covering graphs)\\
        $\mathcal{L}$ & labeled set\\
        $\mathcal{U}$ & unlabeled set\\
        $\mathcal{Q}$ & query set\\
        $n$, $d$ & Number of data points and feature dimension \\
        $C$ & Number of classes \\
        $b$ & Labeling budget \\
        $p$ &Power applied to edge lengths \\
        $\ell_p (x,y,\mathcal{X})$ & $p$-weighted graph Fermat distance between $x$ and $y$\\
        $\epsilon$ & Euclidean ball radius for connecting geometric graph\\
        $r$ & Fermat ball radius for covering graph\\
        
        \hline
    \end{tabular}
    \caption{Notation Table}
    \label{tab:notations}
\end{table}  

\subsection{Fermat Distance}

Discrete Fermat distances can be viewed as classical shortest-path distances in a graph in which edge lengths are weighted according to a penalty parameter $p \in [1, \infty)$. \\

\begin{definition}[Fermat Distance]
\label{sec:fermat_distance}
    For $p \in [1, \infty), x,y \in \mathbb{R}^d$, and some finite set $\mathcal{X} \subset \mathbb R^D$, the (discrete) unrooted $p$-weighted Fermat distance from $x$ to $y$ is:
\begin{equation}
    \ell_p (x,y,\mathcal{X}) = \min_{\pi = \{x_i\}_{i=1}^T} \bigg(\sum_{i=1}^{T-1}\|x_i - x_{i+1}\|^p\bigg),
    \end{equation}
\end{definition}
where $\pi \subset \mathcal{X}$ is a path of dataset points with $x_1 =x$ and $x_T =y$ and $\|\cdot\|$ is the Euclidean norm.


The quantity $\ell_p(\cdot,\cdot,\mathcal{X})$ can be interpreted as the shortest-path distance on the complete graph induced by $\mathcal{X}$, where each edge is assigned the weight $\|x_i-x_j\|^p$. When $p =1$, we recover the Euclidean distance. As $p$ increases, the largest edge lengths along a path contribute heavily to the distance defined in $\eqref{sec:fermat_distance}$. This results in paths through higher-density regions with shorter edge lengths being promoted. When  $p \rightarrow \infty$, 
\[
    \ell_{\infty}(x,y) =\min_{\pi = \{x_i\}_{i=1}^T} \max_{i=1,\cdots T-1}\|x_i -x_{i+1}\|, 
\]
which is called the longest-leg path distance \citep{little2020path}. Outside the extreme values of $p = 1$ and $p = \infty$, $\ell_p$ balances taking a path through regions of high density and taking a ``short'', Euclidean distance path. 

\begin{figure}
    \centering
    \includegraphics[width=\linewidth, height=2.6cm]{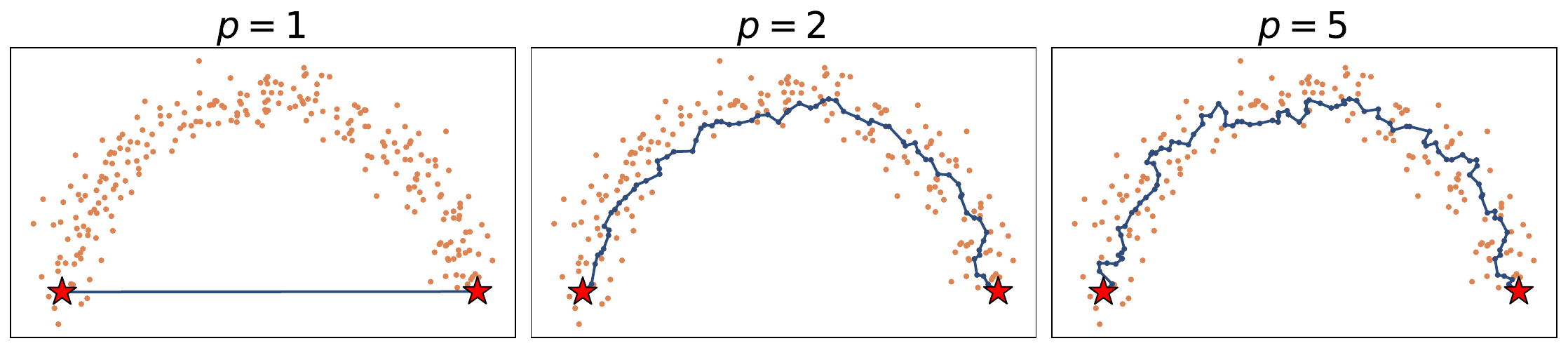}
    \caption{Fermat paths between two points at opposite ends of a moon-shaped dataset for different $p$ values. }
    \label{fig:fermat_path}
\end{figure}

To build intuition for the Fermat distance, Figure~\ref{fig:fermat_path}
shows Fermat paths between two points at opposite ends of a moon-shaped
cluster for $p = 1, 2, 5$. For $p = 1$, the path is the straight segment
between the endpoints, and the Fermat distance reduces to the Euclidean
distance {\color{black}due to the triangle inequality}. For larger $p$, long steps are heavily penalized, so the path
instead takes many short steps through dense regions and follows the shape
of the data distribution.

\begin{figure*}
    \centering
    \includegraphics[width=\linewidth]{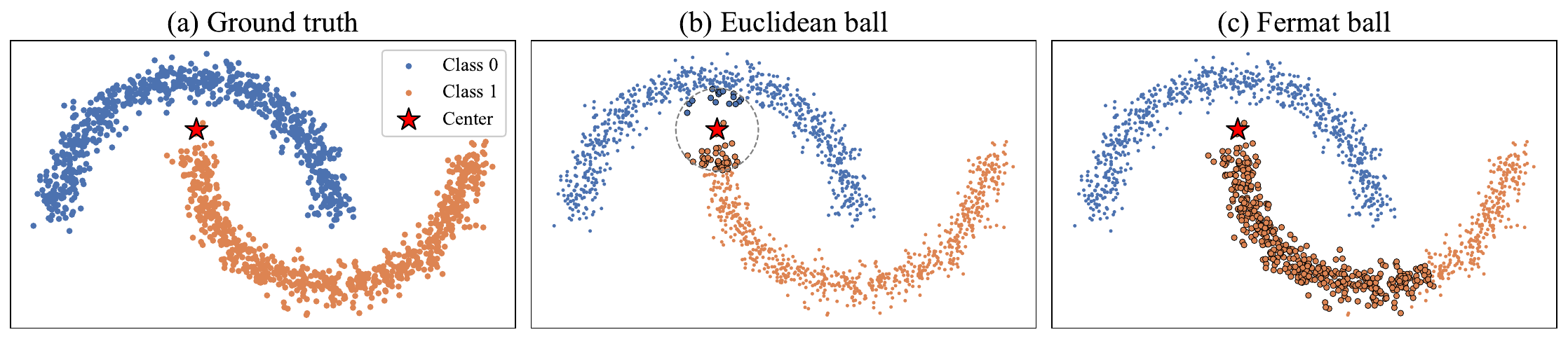}
    \caption{(a) The two-moon dataset with the chosen center (star) at the tip of the lower moon. (b) Euclidean ball with $\epsilon = 0.3$; the dashed circle marks its radius. (c) Fermat ball with $p=3$ and $r = 0.15^p$. The Euclidean ball crosses the gap into the upper moon, while the Fermat ball, despite containing more points, follows the dense region of the lower moon and contains points that are mostly of the same class. }
    \label{fig:toy_balls}
\end{figure*}

\textbf{Toy Example.}
It is important to note here how this change to a data-adaptive metric facilitates our approach. By changing the notion of distance between data points $x, y \in \mathcal{X}$, we capture a fundamentally different geometric structure represented in the dataset. Consider Figure~\ref{fig:toy_balls}, which shows a two-moon dataset consisting of $n = 600$ points in $\mathbb{R}^2$ sampled from two interleaving half-circles, one per class, with Gaussian noise of standard deviation $0.09$. We select a point at the top of the lower moon, which lies inside the arc of the upper moon, and compare its Euclidean ball with a Fermat ball.

The Fermat metric here considers points in a particular moon to be closer to each other than points in different moons that are close in Euclidean distance. In other words, the Fermat ball stretches along the dense region of the lower moon since crossing the sparse gap is expensive under the Fermat distance. This illustrates how the Fermat ball adapts to the shape of the data distribution and captures more interesting clustering geometries than plain Euclidean distances can. We exploit this different geometry to improve the ProbCover approach for coreset selection in active learning.
 
 \subsection{Graph Construction}
\label{graph_construction}

Consider a dataset
$\mathcal{X}=\{x_1,x_2,\ldots,x_n\}\subset\mathbb{R}^d$,
where each point is represented either by its original features or by a learned embedding.
We first construct a weighted $\epsilon$-neighborhood graph $G_\epsilon=(\mathcal{X},\mathcal{E}_\epsilon)$,
where
\[
\mathcal{E}_\epsilon
=
\left\{
(x_i,x_j):
\|x_i-x_j\|_2^p\leq\epsilon^p,\; i\neq j
\right\}.
\]
The parameter $\epsilon>0$ controls the sparsity and connectivity of the graph. A small $\epsilon$ yields a sparse graph that captures local relationships but may result in disconnected components, whereas a large
$\epsilon$ captures broader neighborhood structure at the risk of introducing spurious connections.

In ProbCover \citep{yehuda2022active},
unweighted $G_\epsilon$ serves directly as the coverage graph. In contrast, our proposed method uses weighted $G_\epsilon$ to construct a coverage graph based on Fermat distances. Specifically,
we define the \emph{Fermat coverage graph}
$G_F=(\mathcal{X},\mathcal{E}_F)$, where
\[
\mathcal{E}_F
=
\left\{
(x_i,x_j):
\ell_p(x_i,x_j)\leq r,\; i\neq j
\right\}.
\]
Here, $\ell_p$ denotes the Fermat distance defined in Section~\ref{sec:fermat_distance} and $r>0$ denotes the Fermat coverage radius, measured on the $p$-powered distance scale induced by the edge weights $\|x_i-x_j\|_2^p$. 
The $\epsilon$-neighborhood graph $G_\epsilon$ and the Fermat coverage
graph $G_F$ are closely related. In particular, when
$r\geq\epsilon^p$, every edge of $G_\epsilon$ is also
an edge of $G_F$, and the two graphs have the same
connected components (Lemma~\ref{lemma:fermat_connectivity}
in Appendix).  We describe an efficient procedure for constructing $G_F$ in the following subsection.

\subsection{Efficient Construction of the Fermat Coverage Graph}
\label{sec:computing_fermat}

The Fermat distance can be computed using Dijkstra's algorithm,
but applying it to a complete, weighted Euclidean distance graph over $\mathcal{X}$ is
computationally expensive. Constructing the Fermat
coverage graph $G_F$, however, only requires identifying points within a prescribed Fermat distance threshold based on the parameter $r$. Specifically, for each
$x\in\mathcal{X}$, we consider the Fermat ball
\[
\mathcal{B}_F(x;r) = \{x' \in \mathcal{X} : \ell_p(x, x') \leq r\},
\]
where $r>0$ denotes the Fermat coverage radius, measured
on the $p$-power distance scale.

We exploit this observation in two steps. First, rather than using a (dense) complete, weighted Euclidean distance graph, we perform shortest-path computations on the (sparser) weighted $\epsilon$-neighborhood graph $G_\epsilon$.
Provided that $r\leq\epsilon^p$, this restriction preserves
the Fermat balls exactly, as established in
Lemma~\ref{lemma:fermat_preservation} in the Appendix. {\color{black}Even if $r\geq \eps^p$, we maintain that distances in the geometric neighborhood graph more closely align with distances intrinsic to the dataset than those in the fully connected graph, assuming that $\eps$ is so large that the data sample resolves the data manifold at length scale $\eps$.}

Second, we run Dijkstra's algorithm from each vertex with
a distance cutoff of $r$. Each search terminates once the
smallest tentative distance exceeds this threshold. The
vertices reached within the cutoff form exactly the
Fermat ball $\mathcal{B}_F(x;r)$.

These Fermat balls determine the edge set $\mathcal{E}_F$,
allowing us to construct the Fermat coverage graph without
computing all pairwise Fermat distances.

\section{PROPOSED METHOD}
\label{sec: proposed_method}
Let $\mathcal{X} = \{x_1, x_2, \cdots x_n\}\in \mathbb R^d$ be a dataset, represented either by raw features or by learned embeddings.  The \textit{labeled set} $\mathcal{L}\subseteq \mathcal{X}$ contains the points whose labels $y_i =y(x_i) \in \{1,2,\cdots,C\}$ are known. The \textit{unlabeled set} is  $\mathcal{U} =\mathcal{X}\setminus\mathcal{L}$. Given a budget $b$, the goal is to select a \textit{query set} $\mathcal{Q} \subseteq \mathcal{U}$ (with $|\mathcal{Q}| =b$) whose points will be sent for annotation. 

Our method builds on the ProbCover algorithm~\citep{yehuda2022active},
replacing its Euclidean distance with the Fermat distance. We summarize the
procedure below in three parts.


\paragraph{Graph construction}
We first construct the Fermat coverage graph $G_F$ following
the procedure described in Section~\ref{graph_construction} and Section~\ref{sec:computing_fermat}.
Specifically, we compute the Fermat ball $\mathcal{B}_F(x;r)$ for each
$x\in\mathcal{X}$ using truncated Dijkstra's algorithm on
the weighted $\epsilon$-neighborhood graph $G_\epsilon$.
The coverage graph $G_F$ is then formed by connecting each
point $x$ to all other points in $\mathcal{B}_F(x;r)$.


\paragraph{Sampling}
We initialize the coverage graph by removing all incoming edges to the vertices covered by the labeled set, i.e.\ to every $x' \in \mathcal{B}_F(x;r)$ for $x \in \mathcal{L}$, so that already covered points no longer contribute
to the selection.  We then perform $b$
iterations. In each iteration, we select the vertex $x_{\max}$ with the
highest out-degree for annotation, remove all incoming edges to $x_{\max}$
and its neighboring vertices, and add $x_{\max}$ to the query set
$\mathcal{Q}$. The procedure terminates once the query budget is exhausted, i.e.,\ when $|\mathcal{Q}| = b$.


\paragraph{Prediction}
After the query budget is exhausted, we augment the labeled set with the
queried samples and train a 1-nearest neighbor (1-NN) classifier on the
updated labeled set. We evaluate classification performance using a 1-NN classifier based on Euclidean distance to ensure consistency across all comparison methods. Additionally, Euclidean 1-NN classification is computationally less expensive than a Fermat distance-based 1-NN classifier.  {\color{black} The use of a simple classifier is predicated on the assumption of a data-sparse setting, which does not allow for the training of advanced models like neural networks.}

\begin{figure*}[!tp]
    \centering
    \includegraphics[
        width=\textwidth,
        trim={0.1cm 4.5cm 0.1cm 4.2cm},
        clip
    ]{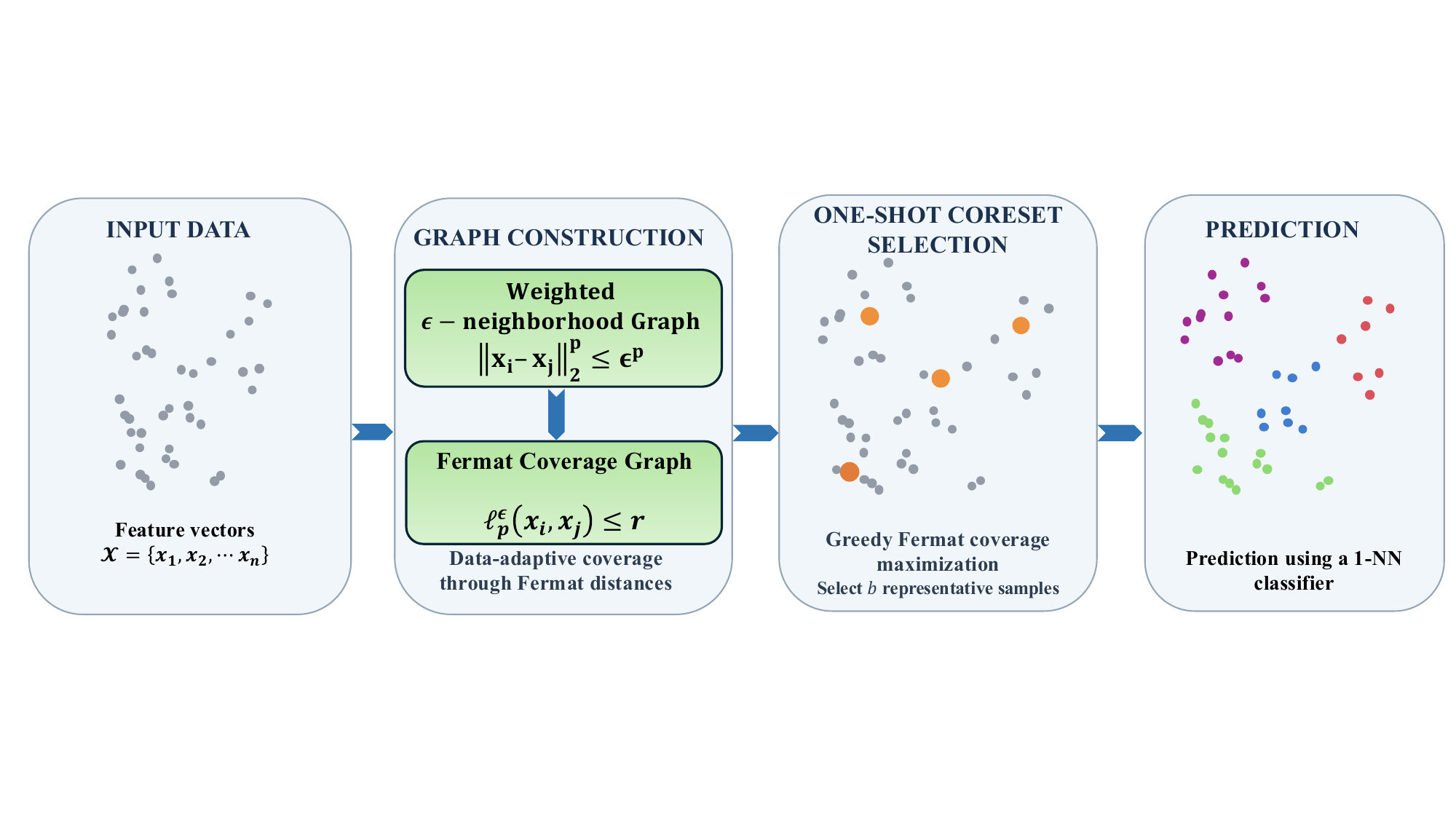}
    \caption{Flowchart of PC-Fermat framework. Given an input dataset, we construct a weighted Euclidean $\epsilon$-neighborhood graph and obtain the
    Fermat coverage graph using shortest-path computations. A greedy coverage-based selection procedure then identifies $b$ representative samples. Finally, predictions are made using a 1-NN classifier, and classification accuracy is evaluated across different labeling budgets.  } 
    \label{fig:pc-fermat}
    \vspace{0.5cm}
\end{figure*}

An overview of the PC-Fermat framework is
illustrated in Figure~\ref{fig:pc-fermat}.

\SetKwFor{ForAll}{for all}{do}{end}
\begin{algorithm}[!htbp]
\caption{ProbCover on Fermat Distance Graph}
\label{alg:fermat-probcover}

\KwIn{
Unlabeled set $\mathcal U$,
labeled set $\mathcal L$,
budget $b$,
Euclidean neighborhood radius $\epsilon$,
Fermat coverage radius $r$,
exponent $p$
}
\KwOut{Query set $\mathcal Q$}

$\mathcal{X} \gets$ learned embeddings or raw features of
$\mathcal{U} \cup \mathcal{L}$\;

\vspace{0.5em}
\tcp{Graph construction}
Construct the $\epsilon$-neighborhood weighted graph
$G_\epsilon=(\mathcal X,\mathcal E_\epsilon)$,
where
$\mathcal E_\epsilon
=\{(x,x'):\|x-x'\|_2^p\leq\epsilon^p,\ x\neq x'\}$,
with edge weights $\|x-x'\|_2^p$\;




\ForAll{$x\in\mathcal X$}{
    $\mathcal B_F(x;r)\gets$
    vertices reached by Dijkstra on the weighted
    $G_\epsilon$ from $x$ with distance cutoff $r$\;
}

Construct the Fermat coverage graph
$G_F=(\mathcal X,\mathcal E_F)$, where
$\mathcal E_F
=\{(x,x'):x'\in\mathcal B_F(x;r),\ x'\neq x\}$\;

\vspace{0.5em}
\tcp{Greedy coverage-based sample selection}
\ForAll{$c \in \mathcal{L}$}{
    Remove incoming edges to vertices covered by $c$:
    
    $\mathcal E_F \gets \mathcal E_F \setminus \left\{(x', x) \in \mathcal E_F : x \in \mathcal{B}_F(c;r)\right\}$\;
}

$\mathcal{Q} \gets \emptyset$\;

\For{$i \gets 1$ \KwTo $b$}{
    $c \gets
    \displaystyle
    \arg\max_{u \in \mathcal{U}}
    \operatorname{outdeg}_{G}(u)$\;

    $\mathcal{Q} \gets \mathcal{Q} \cup \{c\}$\;

    Remove incoming edges to vertices covered by $c$: \;

    $\mathcal E_F \gets
    \mathcal E_F \setminus
    \left\{
    (x',x) \in \mathcal E_F :
    (c,x) \in \mathcal E_F
    \right\}$\;
}

\Return{$\mathcal{Q}$}\;
\end{algorithm}

\section{ Empirical Results} \label{sec: results}
In this section, we report a set of empirical results, comparing our method with other active learning strategies. First, we detail the datasets used in the experiments.  
\subsection{Datasets}

\begin{itemize}
\item The COIL-20 dataset \citep{nene1996columbia} contains 1440 normalized images of 20 objects (i.e., classes). The objects were placed on a motorized turntable. With a fixed camera, the turntable was rotated through 360 degrees to capture different poses every 5 degrees. There are a total of 72 images per object. The dataset consists of 1,440 data points and has 1,024 attributes.
\vspace{0.15cm}
\item The Opt-Digits dataset \citep{optdigits} is a multiclass dataset of 5,620 handwritten digits. Each handwritten digit is recorded as an 8 $\times$ 8 matrix. Thus, each element has 64 attributes. There are 10 classes.
\vspace{0.15cm}
\item The Landsat dataset \citep{landsat} contains 6,435 elements, each with multi-spectral values of pixels in 3x3 neighborhoods in a satellite image. For this dataset, there are six classes, and the classification of an element is derived from the ground truth label of the central pixel. Each of the data elements contains the pixel values in the four spectral bands of each pixel in the 3×3 neighborhood. Therefore, each feature vector consists of 36 values. The six classes are red soil, cotton crop, gray soil, damp gray soil, soil with vegetation stubble, and very damp gray soil. 
\vspace{0.15cm}
\item The USPS dataset \citep{usps} contains 9,298 grayscale images of handwritten digits, each with a resolution of $16\times16$ pixels. The images were obtained by automatically scanning envelopes by the United States Postal Service in Buffalo, New York. The grayscale images are centered and normalized and show various font styles. There are 10 classes; each element is an image of a handwritten digit ``0" through ``9". 
\vspace{0.15cm}
\item The Jasper Ridge dataset \citep{jasper} is widely used for hyperspectral image classification. The original dataset consists of a $512 \times 614$ pixel image with 224 spectral channels. For our experiments, we consider a $100 \times 100$ subimage spanning 198 channels. Due to dense water vapor and atmospheric effects, the channels 1--3, 108--112, 154--166 and 220--224  are removed.
\vspace{0.15cm}
\item The Pendigits dataset \citep{pen-based_recognition_of_handwritten_digits_81} is a collection of pen-based digit samples. It has a total of 10,992 samples with 16 features and 10 target classes (digits 0-9). The database is created by collecting 250 samples from 44 writers. 
\vspace{0.15cm}

\item The MNIST dataset \citep{lecun2010mnist} is a handwritten digit classification dataset consisting of 70,000 grayscale images of handwritten digits from 0 to 9. Each image has a resolution of $28 \times 28$ pixels and is flattened into a 784-dimensional feature vector. We use a variational autoencoder (VAE) \citep{kingma2013auto} to learn a lower-dimensional latent representation of the images, which is then used as the embedding for our experiments. Specifically, each 784-dimensional input is mapped to a 20-dimensional latent representation.

\item \textcolor{black}{The Fashion-MNIST dataset \citep{xiao2017fashion} consists of $70,000$ images with $10$ classes. Each image has a 784-dimensional feature vector. Similar to MNIST,  we use a variational autoencoder (VAE) \citep{kingma2013auto} to learn a lower-dimensional latent representation of the images, which is then used as the embedding for our experiments. Specifically, each 784-dimensional input is mapped to a 30-dimensional latent representation.}
\end{itemize}

The details of the datasets are outlined in Table \ref{tab:datasets}.


\begin{table}[t!]
    \centering
    \caption{Datasets used in the experiments.}
    \label{tab:datasets}
    \vspace{0.05cm}
    \begin{tabular}{|p{2cm}||p{1cm}|p{1cm}|p{1cm}|p{0.9cm}|}
        \hline
        Dataset & Samples & Features & Type & Classes\\
        \hline
        Coil-20     & 1,440  & 1,024 & Raw       & 20 \\
        Opt-Digits  & 5,620  & 64    & Raw       & 10 \\
        Landsat     & 6,435  & 36    & Raw       & 6 \\
        USPS        & 9,298  & 256   & Raw       & 10 \\
        Jasper      & 10,000 & 198   & Raw       & 4 \\
        Pendigits   & 10,992 & 16    & Raw       & 10 \\
        MNIST       & 70,000 & 20    & Embedd. & 10 \\
        Fashion-MNIST       & 70,000 & 30    & Embedd. & 10 \\
        \hline
    \end{tabular}
\end{table}

\subsection{Comparison to Other Methods}

\begin{figure*}[t!]
\centering 
\begin{subfigure}(a)  
\centering
\includegraphics[scale=.40]{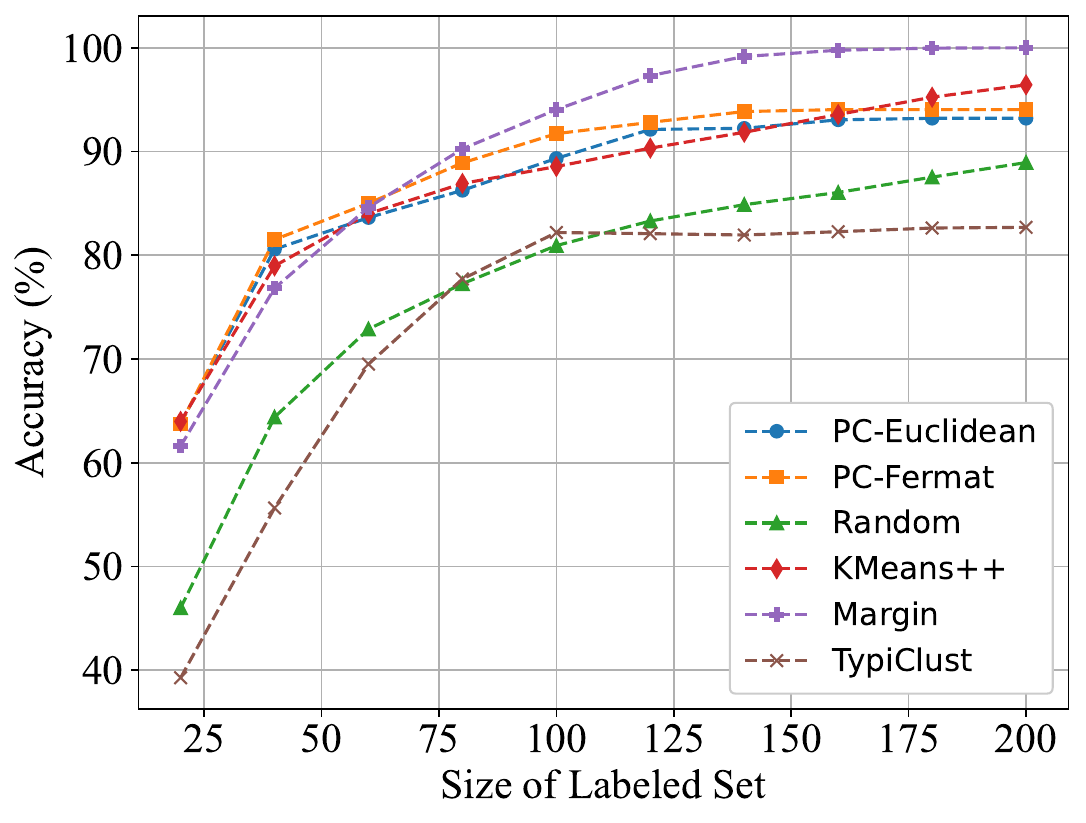} 
\label{coil}
\end{subfigure}
\begin{subfigure}(b)  
\centering
\includegraphics[scale=.40]{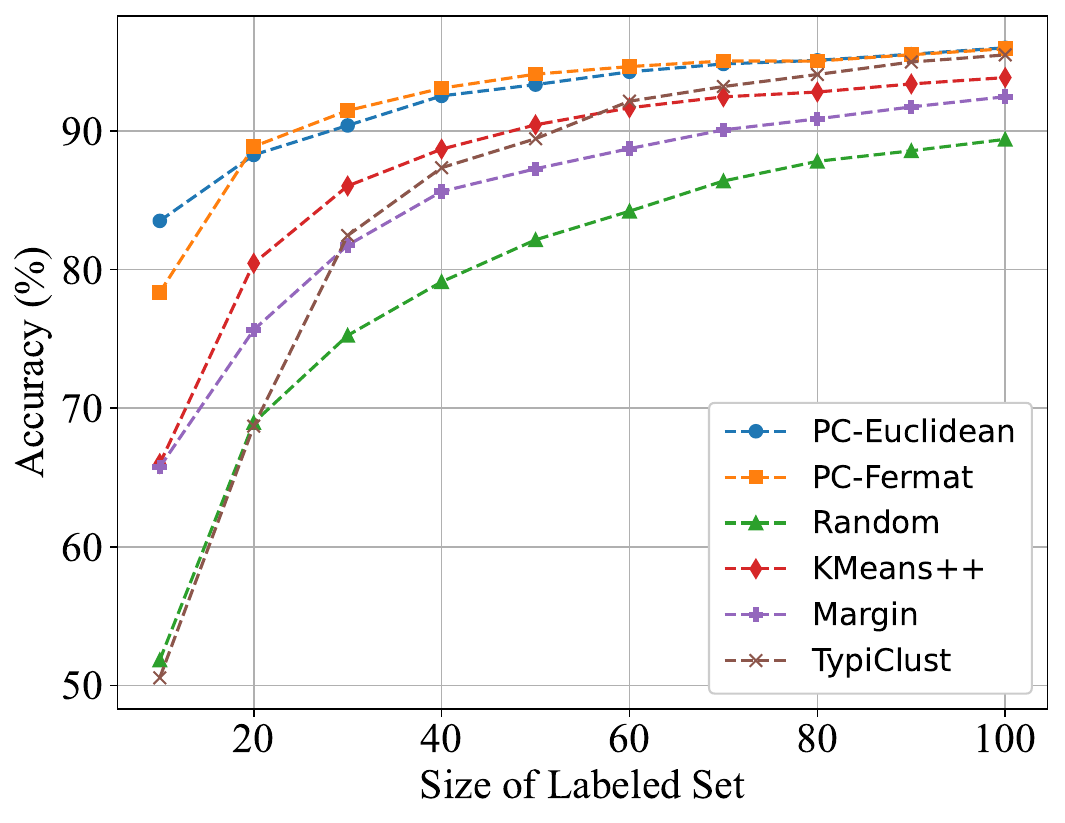} 
\label{opt}
\end{subfigure} 
 
\begin{subfigure}(c)  
\centering
\includegraphics[scale=.40]{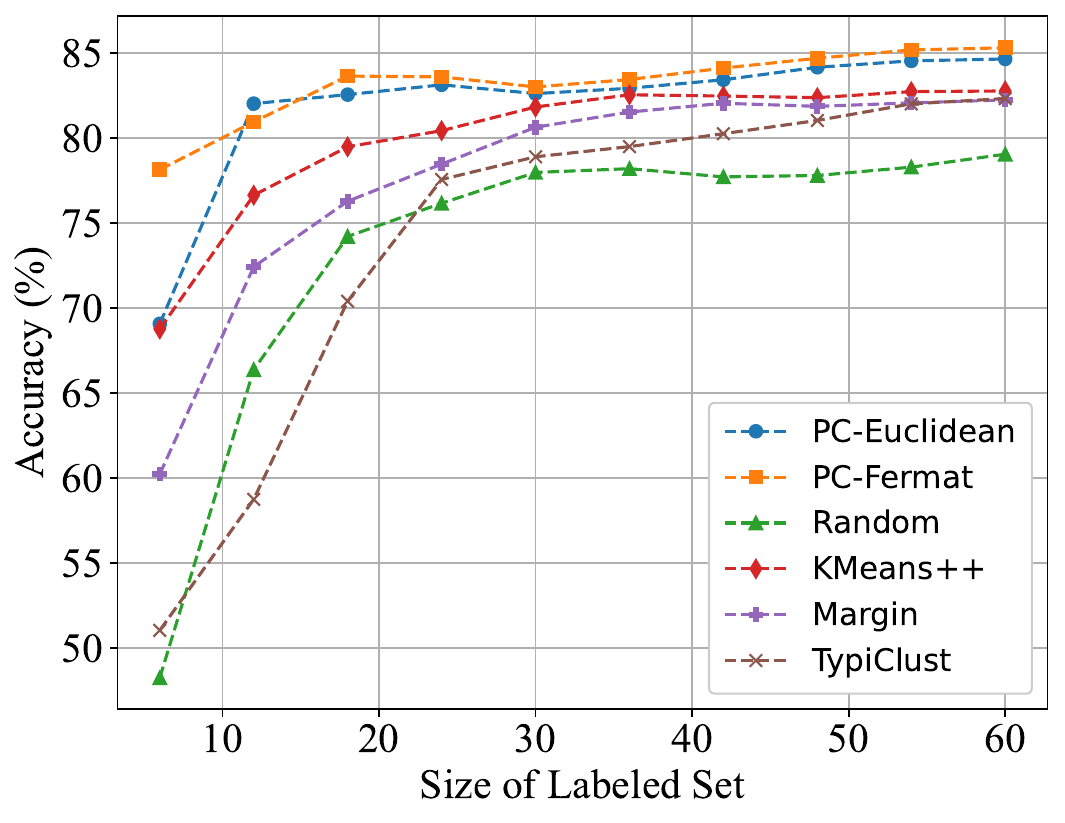} 
\label{usps}
\end{subfigure}
\begin{subfigure}(d)  
\centering
\includegraphics[scale=.40]{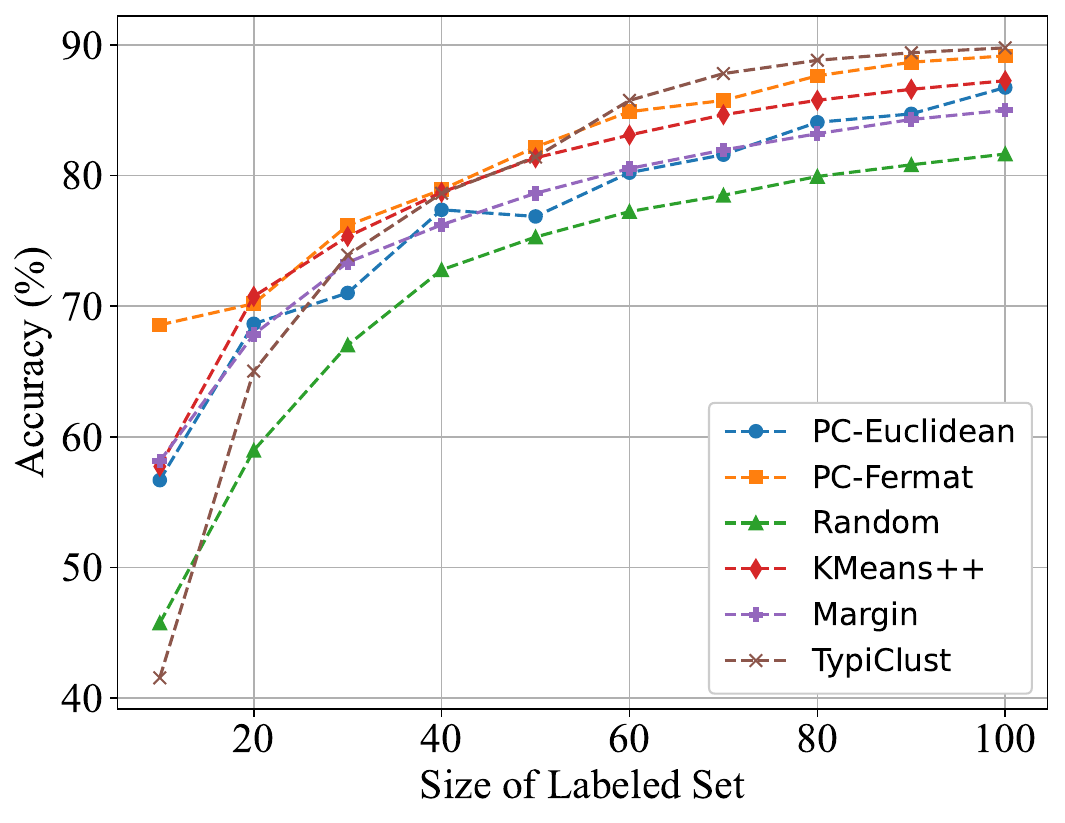} 
\label{landsat}
\end{subfigure}

\begin{subfigure}(e)  
\centering
\includegraphics[scale=.40]{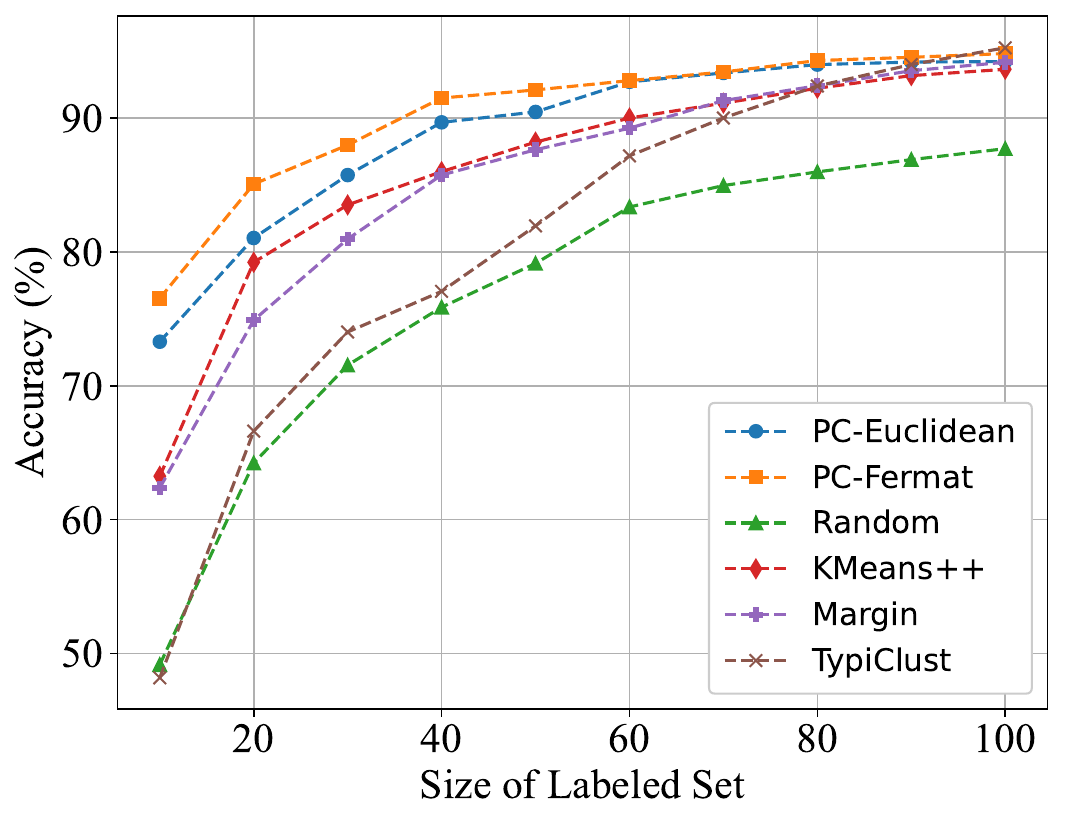} 
\label{usps}
\end{subfigure}
\begin{subfigure}(f)  
\centering
\includegraphics[scale=.40]{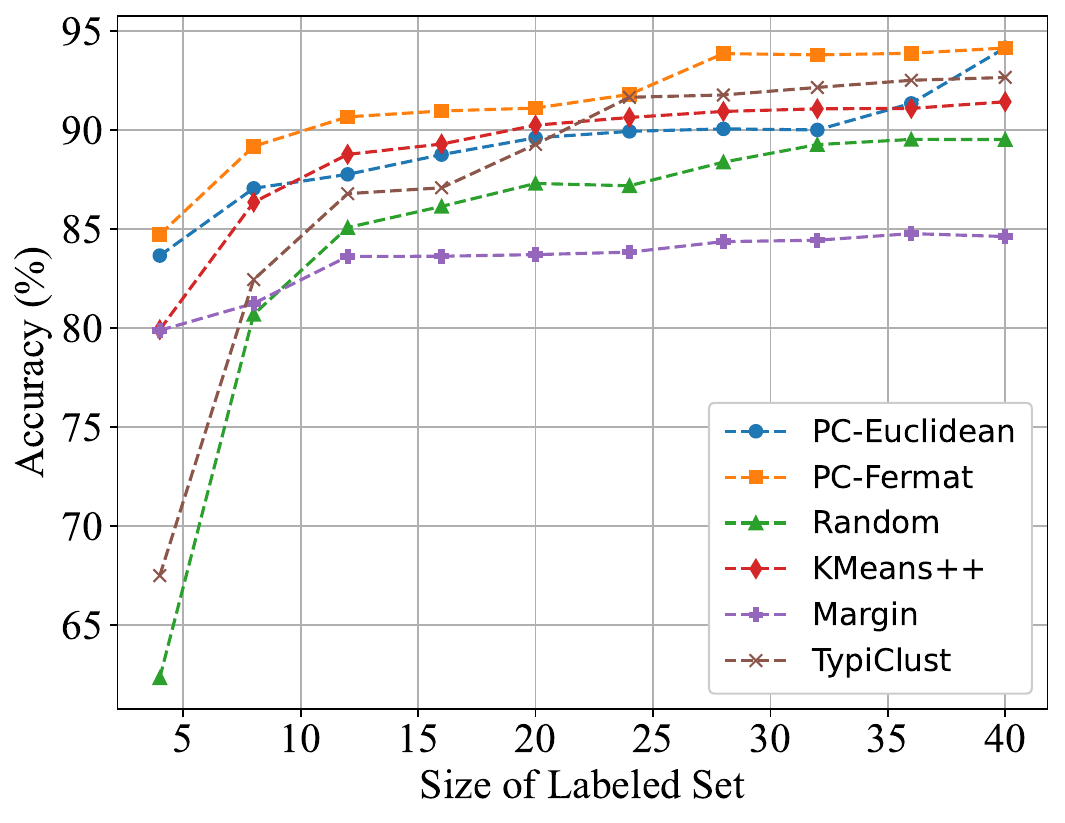} 
\label{usps}
\end{subfigure}

\begin{subfigure}(g)  
\centering
\includegraphics[scale=.40]{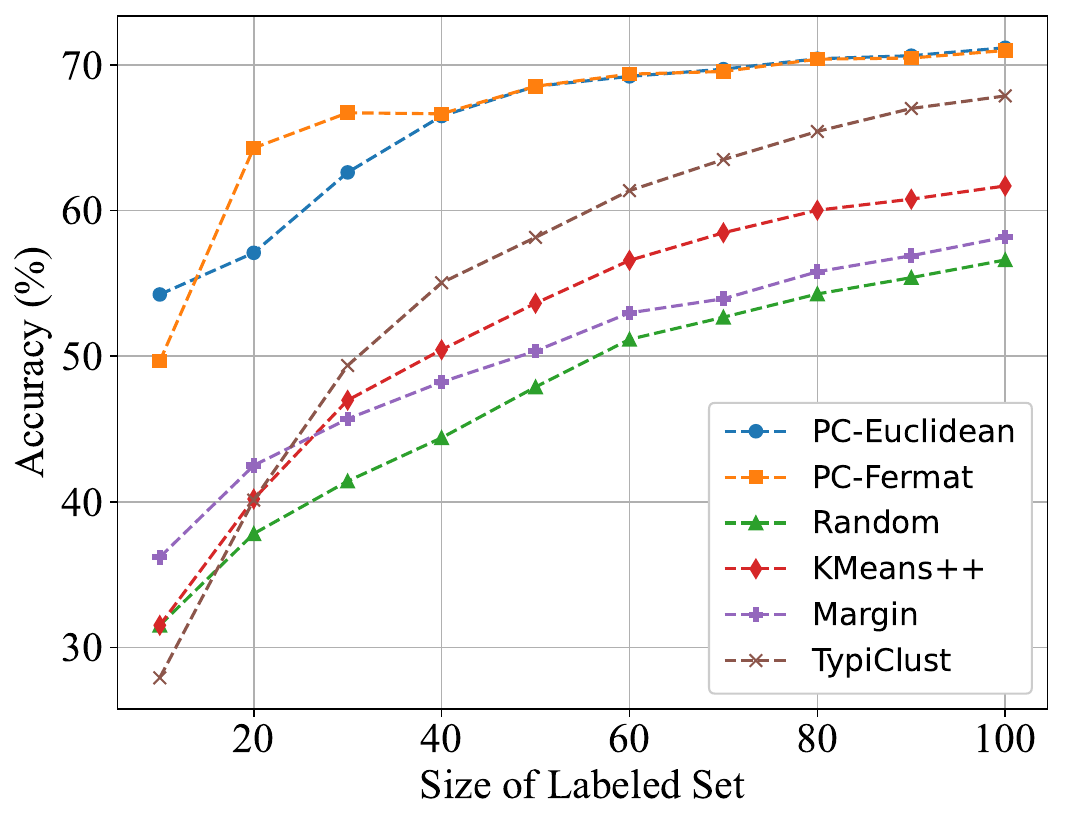} 
\label{usps}
\end{subfigure}
\begin{subfigure}(h)  
\centering
\includegraphics[scale=.40]{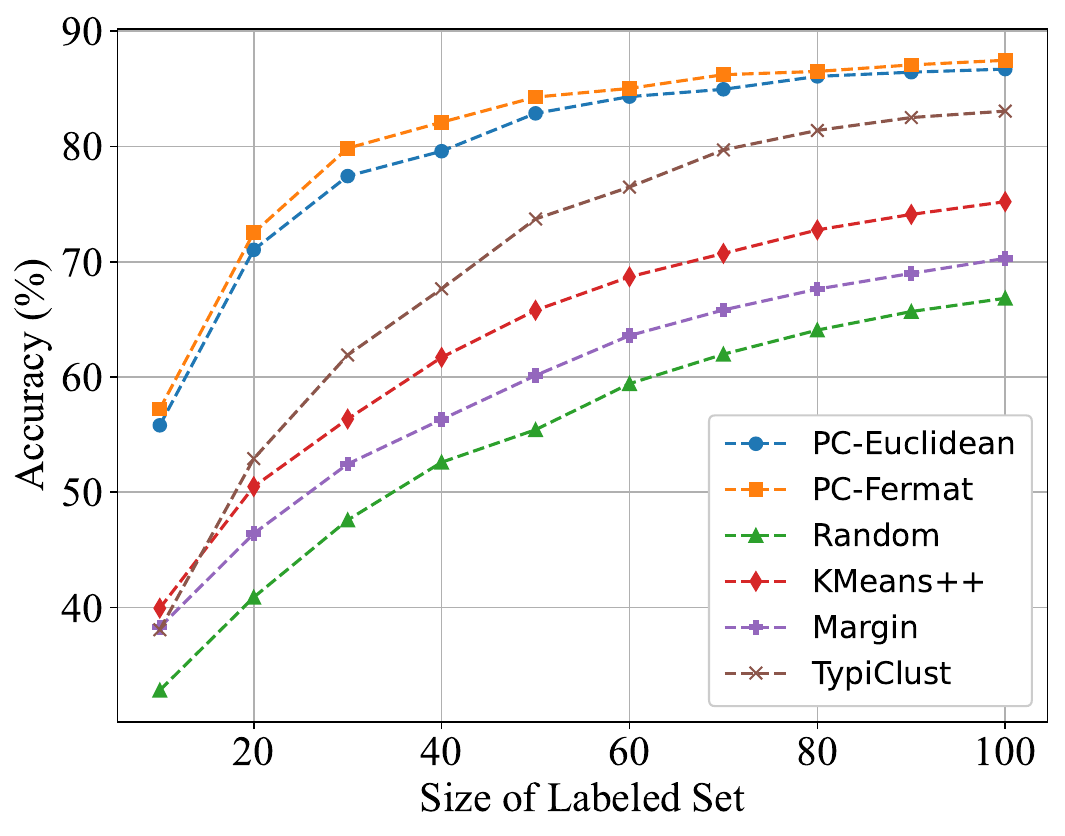} 
\label{usps}
\end{subfigure}
\caption{Accuracy vs. \# of labeled points for each dataset: (a) COIL-20  (b) Opt-Digits (c) Landsat (d) USPS
 (e) Pendigits (f) Jasper (g) Fashion-MNIST (h) MNIST. } 
\label{fig:plots}
\end{figure*}

In this section, we present empirical results comparing our proposed method, ProbCover with a Fermat distance graph (hereafter referred to as PC-Fermat), with several existing
active learning methods in the very low-budget regime. In particular, we compare PC-Fermat against Random Sampling, Margin Sampling, KMeans++ \citep{arthur2007k}, ProbCover \citep{yehuda2022active} (hereafter referred to as PC-Euclidean),
and TypiClust \citep{hacohen2022active}. The results of the experiments are shown in Figure \ref{fig:plots}. For a fair comparison, we use a 1-Nearest Neighbor (NN) classifier as the underlying classifier model for all active learning strategies. The 1-NN classifier uses Euclidean distance for all methods. 

Next, we give a brief overview of our comparison methods. In particular, Random selects unlabeled samples uniformly at random. Margin is an uncertainty-based method that queries samples with the smallest prediction margin between the top two classes. Moreover, KMeans++ \citep{arthur2007k} is a seeding algorithm that starts by first selecting one point at random, then iteratively choosing subsequent centers with probabilities proportional to their squared distance from the nearest existing center. This iterative process leads to more diverse and effective seeding than random initialization. ProbCover \citep{yehuda2022active} (PC-Euclidean) is an active learning method for low-budget regimes and it seeks to maximize probability coverage. Lastly, TypiClust \citep{hacohen2022active} clusters the feature space for diversity and queries the typical examples in each cluster that favor the selection of representative examples.

For the implementation of ProbCover (PC-Euclidean) and TypiClust, we utilize the code obtained from the authors' repository. For KMeans++, we use the implementation available in the \texttt{scikit-learn} Python library \citep{scikit-learn}. Our Fermat ProbCover (PC-Fermat) implementation is based on the original ProbCover codebase with the necessary modifications to support the Fermat distance computation. 

\subsection{Hyperparameter Selection}
In this section, we describe the hyperparameter selection procedure used in our experiments. One of the important hyperparameters of our method is the radius of the {\color{black}(Euclidean or Fermat) ball for ProbCover}, denoted by $r$. The parameter $r$ directly controls the trade-off between coverage and local purity. A larger value of $r$ provides higher coverage but results in less pure neighborhoods. In addition, most of the datasets considered in our experiments do not have pre-computed embeddings. Therefore, we carefully tune $r$ for each dataset and report the selected values in the supplementary materials.


{\color{black}In the case of PC-Fermat with Fermat power $p$, we couple $r=\eps^p$ in light of Lemma \ref{lemma:fermat_preservation} and thus seek a single parameter $\eps$ for which the majority of data points are within a few large connected components to ensure that the geometry of the data clusters is well-resolved. We illustrate a typical and a non-typical situation in the growth of large connected components in Figure \ref{fig:connected_component_size} as $\eps$ increases using the COIL-20 and Landsat datasets, respectively.}

\subsection{ Performance and Discussion}
\label{discussion}
For each dataset, all methods except Margin sampling are initialized with an empty labeled set.  Given a labeling budget $b$, we query $b$ data points for their labels. For Margin sampling, we initialize the labeled set randomly with one label per class. Our proposed method PC-Fermat uses Fermat distance for graph construction, while PC-Euclidean uses Euclidean distance for graph construction.  Figure \ref{fig:plots} reports the classification accuracy of all evaluated methods across the datasets. 

\begin{figure*}[t!]
\centering 
\begin{subfigure}(a)  
\centering
\includegraphics[scale=.35]{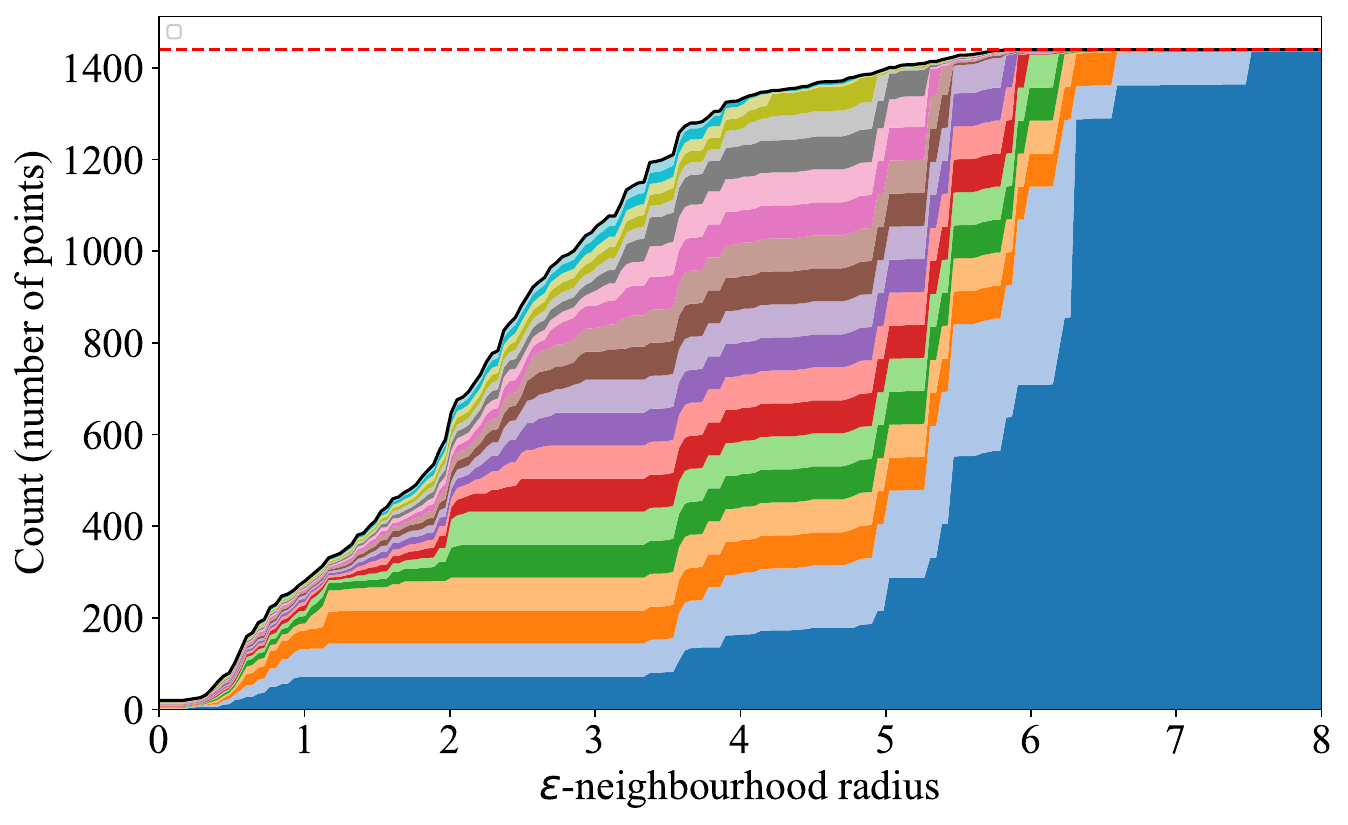} 
\label{usps}
\end{subfigure}
\begin{subfigure}(b)  
\centering
\includegraphics[scale=.35]{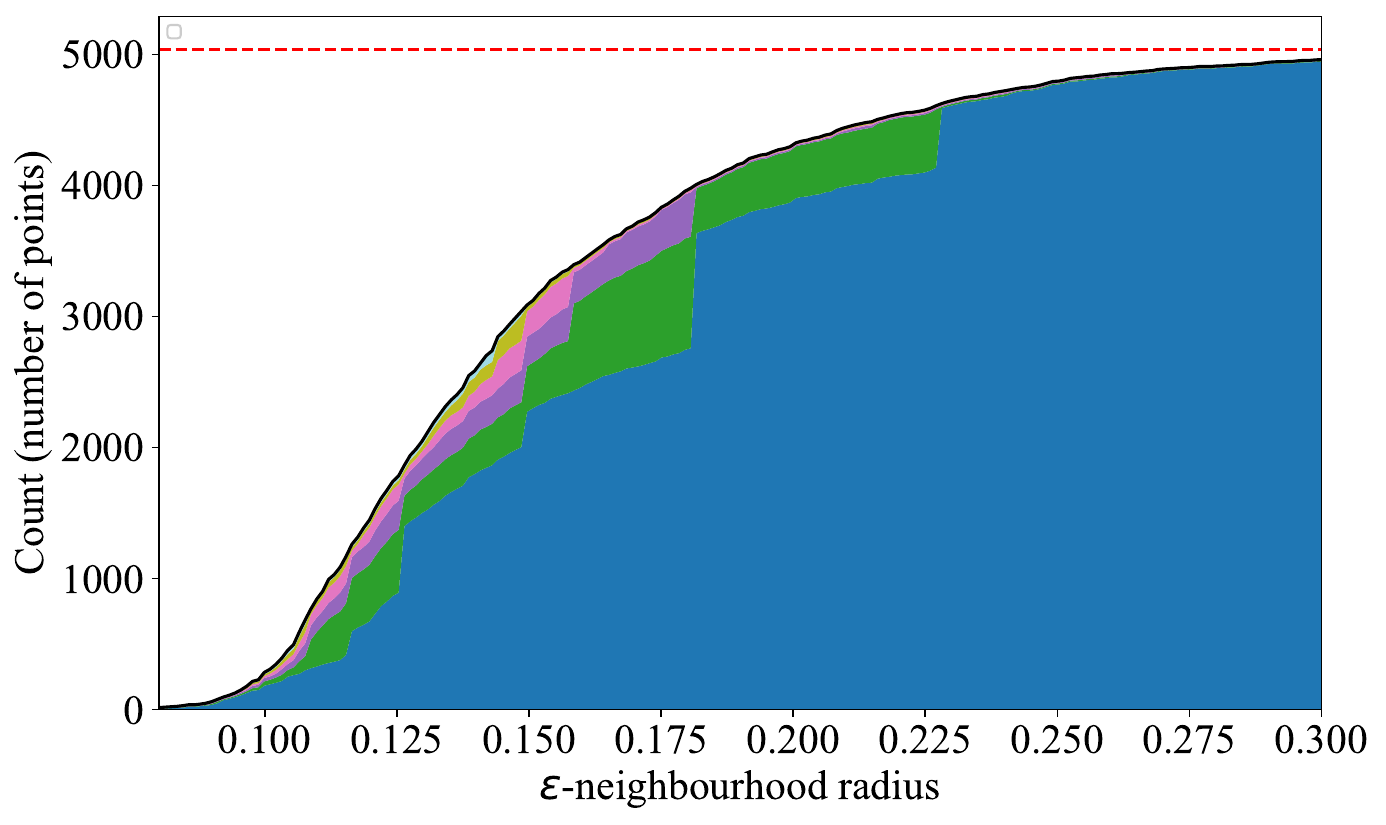} 
\label{usps}
\end{subfigure}
\caption{Evolution of largest connected components of the $\epsilon$-neighborhood graph with increasing $\epsilon $. Each colored band shows the size of one component; the black line shows their total size, and the red dashed line marks the total number of points. COIL-20 (left) contains several well-separated clusters, while Landsat (right) is dominated by a single connected component.  }
\label{fig:connected_component_size}
\end{figure*}

Overall, our proposed PC-Fermat achieves the most competitive performance across most datasets, outperforming the random sampling baseline and generally improving upon the competing methods. {\color{black} We observe in Figure \ref{fig:connected_component_size} that the $\eps$-neighborhood graph for COIL-20 has multiple connected components of large comparable sizes for a wide range of values of $\eps$, suggesting a set of several well-separated clusters. In this specific setting, Margin outperforms our PC-Fermat and all other representation-based methods, suggesting an early transition from an exploration-dominated to an exploitation-dominated regime. In the more typical case of the Landsat dataset (Figure \ref{fig:connected_component_size}, right), data cannot be separated as neatly into Euclidean clusters.}






\section{Conclusion}
\label{sec: conclusion}
In this paper, we introduce Fermat ProbCover (PC-Fermat), a coverage-based active learning method that integrates the Fermat distance, a density-aware metric, into the ProbCover framework.  The proposed method is well suited to the low-budget regime, where only a few labeled examples are available, a common scenario for many applications since manual data labeling is frequently both time-consuming and cost-prohibitive. Experimental results on eight classification datasets under varying labeling budgets show that our method outperforms {\color{black}the Euclidean variant of} ProbCover and other active learning methods on most of the datasets. 

{\color{black}In terms of computational cost, random sampling and $k$-means++ are significantly more lightweight algorithms than ProbCover, and PC-Fermat increases the computation cost further by requiring intrinsic distances to be computed, at least up to the threshold value of $r=\eps^p$, at which point Dijkstra's algorithm terminates early. However, in the examples considered above, it is the most consistently competitive algorithm and is recommended when the cost of data annotation vastly exceeds that of finding a representative subset; this is often the case in practical applications.}

\section{Acknowledgments}
The work of EM was partly supported by grant NSF DMS-2512644.
The work of DD and SW was partly supported by grant NSF DMS-2424801.




\bibliographystyle{IEEEtran}
\bibliography{IEEEabrv,references}

\appendix 


In this section, we establish the two properties of the Fermat coverage graph. The first result characterizes its relationship
with the underlying Euclidean neighborhood graph in terms of edge inclusion and connectivity. The second result provides conditions under which computing Fermat distances on the
Euclidean neighborhood graph preserves the Fermat distance computed from the complete graph.

 \begin{lemma}
 \label{lemma:fermat_connectivity}
If we construct an $\eps$-neighborhood graph for the geometry graph and then build the coverage graph based on pairwise $p$-Fermat distances in the geometry graph, then if $r\geq\eps^p$
\begin{itemize}
\item every edge of the geometry graph is also an edge of the coverage graph and
\item both graphs have the same connected components.
\end{itemize}
\end{lemma}
\begin{proof}
    Let $G_{\epsilon}=(\mathcal{X},\mathcal{E}_{\epsilon})$ be the $\eps$-neighborhood graph for the geometry graph and  let $G_{F}=(\mathcal{X},\mathcal{E}_F)$ be the coverage graph based on $p$-Fermat distances on $G_F$. Consider any edge $(x,y) \in \mathcal{E}_\epsilon$. By definition of $\eps$-neighborhood graph, the distance between the vertices $x$ and $y$ satisfies
    \[d(x,y)\leq \epsilon.\]

For $p \in [1, \infty),$ and vertices $ x,y \in \mathcal{X}$, let  $\ell_p^\epsilon (x,y)$ denote the $p$-Fermat distance between the vertices $x$ and $y$ over some path. If $r\geq \eps^p$  we have the following relation

\[\ell_p^\epsilon (x,y)  \leq d(x, y )^p \leq \epsilon^p \leq r.\]
This gives $\ell_p^\epsilon (x,y) \leq r$. So by the definition of a neighborhood graph, this implies that $(x,y) \in \mathcal{E}_{F}$. Hence, every edge of the geometry graph is also an edge of the coverage graph. 

Because $\mathcal{E}_{\epsilon}\subseteq \mathcal{E}_{F}$, any path in $G_{\epsilon}$ is also a path in $G_{F}$. This means each connected component of the geometry graph is contained in a connected component of the coverage graph. It remains to show that the coverage graph does not merge two distinct connected components of the geometry graph. 

To this end, consider an edge $(x,y) \in \mathcal{E}_{F}$. By definition of the coverage graph, there exists a path in $G_{\epsilon}$ connecting $x$ and $y$ with $\ell_p^\epsilon \leq r$. In particular, such a path certifies that $x$ and $y$ lie in the same connected component of $G_{\epsilon}$. Hence, there is no edge in the coverage graph that connects vertices from two different components of the geometry graph.  Therefore, both graphs have the same connected components. 

\end{proof}

 \begin{lemma}
 \label{lemma:fermat_preservation}
Let $\epsilon>0$, $p>1$, and $0<r\leq\epsilon^p$.
Denote by $\ell_p$ and $\ell_p^\epsilon$ the Fermat distances
computed on the complete weighted graph and the weighted
$\epsilon$-neighborhood graph $G_\epsilon$, respectively.
Then, for any $x,y\in\mathcal{X}$,
\[
\ell_p(x,y)\leq r
\quad\Longleftrightarrow\quad
\ell_p^\epsilon(x,y)\leq r.
\]
Furthermore, whenever either inequality holds,
\[
\ell_p(x,y)=\ell_p^\epsilon(x,y).
\]
 
 \end{lemma}

\begin{proof}
 
Since $G_\epsilon$ is a subgraph of the complete graph,
we have
\[
\ell_p(x,y)\leq \ell_p^\epsilon(x,y).
\]
Thus, if $\ell_p^\epsilon(x,y)\leq r$, then
$\ell_p(x,y)\leq r$.

Conversely, suppose $\ell_p(x,y)\leq r$, and let
$\pi^*$ be a shortest path between $x$ and $y$
in the complete weighted graph. Then
\[
\sum_{(x_u,x_v)\in\pi^*}
\|x_u-x_v\|_2^p
\leq r.
\]
Since all edge weights are nonnegative, every edge
along $\pi^*$ satisfies
\[
\|x_u-x_v\|_2^p
\leq r
\leq\epsilon^p.
\]
Since $p>1$, it follows that
$\|x_u-x_v\|_2\leq\epsilon$.
Thus, every edge of $\pi^*$ belongs to $G_\epsilon$,
and hence
\[
\ell_p^\epsilon(x,x)
\leq \ell_p(x,x).
\]
Combining the two inequalities yields
\[
\ell_p^\epsilon(x,y)=\ell_p(x,y)\leq r,
\]
which completes the proof.
\end{proof}

\end{document}